# Determining Image similarity with Quasi Euclidean Metric

Vibhor Singh[1], Vishesh Devgan[2], Ishu Anand[3]

[1] Department of Computer Science and Engineering Bharati Vidyapeeths College of Engineering, New Delhi, India
vibhor97singh@gmail.com

[2] Department of Computer Science and Engineering Bharati Vidyapeeths College of Engineering New Delhi, India
vdevgan100@gmail.com

[3] Department of Computer Science and Engineering Bharati Vidyapeeths College of Engineering New Delhi, India
ianand.ishu@gmail.com

**Abstract.** Image similarity is a core concept in Image Analysis due to its extensive application in computer vision, image processing, and pattern recognition. The objective of our study is to evaluate Quasi Euclidean metric as an image similarity measure and analyze how it fares against the existing standard ways like SSIM and Euclidean metric. In this paper, we analyzed the similarity between two images from our own novice dataset and assessed its performance against the Euclidean distance metric and SSIM. We also present experimental results along with evidence indicating that our proposed implementation when applied to our novice dataset, furnished different results than standard metrics in terms of effectiveness and accuracy. In some cases, our methodology projected remarkable performance and it is also interesting to note that our implementation proves to be a step ahead in recognizing similarity when compared to

## 1 INTRODUCTION

Similarity measure can be elucidated as the measure of how similar two data/signal are. Alternatively, a distance metric can be termed as similarity measure[9]. Similarity measure using metrics in the context of images, is any distance with dimensions representing features or pixels of an image. If this distance is minute, there will be a higher measure of similarity and as the distance increases, lower will be the measure of similarity. The term similarity between two objects is subjective to the context it's being referred to. It is highly

For example, two food items can be similar in terms of either their color, taste or even calories.

The current section gives a high-level introduction about similarity measures. Section 2 describes the distance metrics used in our research along with their formulas. Section 3 contains the methodology that has been used to achieve our results. Section 4 contains information on how the experiment was setup and the results that were obtained. Section 5 contains the conclusion that the paper draws from the results obtained and finally Section 6 mentions how the results drawn from this paper can be extended further.

## 2  Background

### 2.1  Euclidean Distance

Euclidean Distance is a metric that is widely used to measure distance between two points in Euclidean Space [1]. It is most frequently used when the data being worked upon is continuous in nature. It is often described as the length of the straight line between any two given points. In this paper, we have used Normalized Euclidean Distance.

$$Euclidean((x_1, y_1), (x_2, y_2)) = \sqrt{(x_2 - x_1)^2 + (y_2 - y_1)^2} \tag{1}$$

### 2.2  Structural Similarity Index

The SSIM method is a well-known quality metric used to measure the similarity between two images[4] and is a part of Perceptual Image Processing[10]. SSIM works by extracting structural information of the object which is present in the image. This method was developed by *Wang et al* [2]. The SSIM is calculated for each overlapped image block by using a pixel-by-pixel sliding window, and therefore, it can provide the distortion/similarity map in the pixel domain[8]. It models any distortion in the image as the blend of three constituent factors which are luminance distortion, loss of correlation and contrast distortion.

$$SSIM(x, y) = \frac{(2\mu_x\mu_y + C_1) + (2\sigma_{xy} + C_2)}{(\mu_x^2 + \mu_y^2 + C_1)(\sigma_x^2 + \sigma_y^2 + C_2)} \tag{2}$$

### 2.3  Quasi-Euclidean

Quasi-Euclidean metric, like the name suggests, is similar to Euclidean metric but it differs as it measures the total Euclidean distance along a set of horizontal, vertical and diagonal line segments whereas Euclidean distance is measured only along a straight line [3].

It is a path-based distance metric. Path-based distance metric functions are often used on the discrete plane to compute the distance between two pixels as the length of the shortest (not necessarily unique) path linking them. The degree of approximation to the Euclidean distance depends essentially upon which are the unit moves permitted along the path, and on the weights used to measure them [6].

As Euclidean Distance measures the shortest distance between 2 points, other path-based metrics like quasi-euclidean, try to approximate their value as close to the minimum value as they can. The Quasi-Euclidean metric, unlike other conventional metrics, uses knight moves to approximate to Euclidean distance. The shape formed It is a hexadecagonal metric (Figure 1) as it uses a 5 by 5 neighborhood to approximate to Euclidean distance.

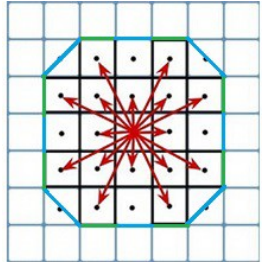

**Fig. 1.** The outline of 5x5 neighborhood gives a hexadecagon outline

$$QE((x_1, y_1), (x_2, y_2)) = |x_1 - x_2| + (\sqrt{2} - 1)|y_1 - y_2| \quad (3)$$
$$\text{if } |x_1 - x_2| > |y_1 - y_2|$$
$$\text{otherwise,}$$
$$QE((x_1, y_1), (x_2, y_2)) = (\sqrt{2} - 1)|x_1 - x_2| + |y_1 - y_2| \quad (4)$$

where $(x_1, y_1)$ and $(x_2, y_2)$ denotes pixel of *image*1 and *image*2 respectively.

## 3  Proposed Methodology

This paper implements Quasi Euclidean metric because this metric takes into account diagonal, vertical as well as horizontal pixels in it's formula while comparing and measuring similarity between two images.

The approach for successful execution of our method is -:

1. First, Anaconda Cloud software[5] and Python (version 3) were set up.
2. Jupyter notebook was employed to write the code and run.
3. Libraries including Sci-kit, Sklearn, Numpy, Scipy, cv2 (OpenCV) and Math were used.

4. The dataset this paper works upon was a novice dataset that was handpicked in a non-deterministic fashion having images categorized into several
5. objects. Two RGB JPEG images of size 800 x 800 were picked at a time and used as input.
6. Next, these RGB images were converted to Gray-scale color palette for simpler calculation and implementation.
7. Now, the two Gray-Scale images were compared with each other by calculating their resolution and then running a loop until maximum values of resolution inside which the Quasi-Euclidean distance was calculated.
8. In this paper, to calculate this metric, its formulae given by (3) and (4) were used.
9. Then, SSIM and Euclidean distances were calculated using pre-defined library functions in Sci-kit.
10. Finally, similarity between the two input images using Quasi-Euclidean and other metrics were obtained and was given as the program's output.

## 4    Application and Experimental Results

Our major intention of this implementation was to extensively verify and validate the accuracy and effectiveness of Quasi-Euclidean distance metric and contrast it against the commonly used metrics: Structural Similarity index (SSIM) and Euclidean Distance metric.

### 4.1    Setup

For our implementation, we constituted a novice dataset that was handpicked in a non-deterministic fashion. The images were scoured from Google resources.

These images were carefully selected as they form an integral part of our implementation. We payed attention that input images were of same size and object of interest inside these images had roughly the same alignment in both the images that had to be compared for consistent results.

On execution of Quasi-Euclidean metric along with SSIM and Euclidean metric assisted us in finding how our novice implementation functions when compared to the standard methods.

Figure 2 illustrates the similarity measures obtained using each method as observed on some of the images. The likelihood of success that our novel methodology will exhibit in the use case of an image retrieval system to assess the viability of our proposed method formed a crucial benchmark for inspection of the results.

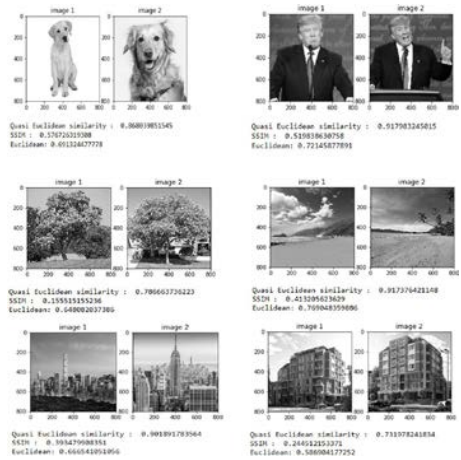

**Fig. 2.** A sample of Similarity measures for each category

### 4.2 Results

Similarity was calculated for six different objects, with all of them making up our novice dataset. These were - Beaches, Cities, Dogs, Housing, People and Trees. Our dataset had 20 pictures belonging to each of the above categories. The average similarity of objects that were actually similar is shown in Table 1.

Quasi-Euclidean metric and Euclidean metric are normalized and can have any value between 0 to 1 whereas, SSIM can have any value between -1 and 1. Higher the value, greater is the similarity in-case of all metrics.

**Table 1.** Average similarity scores of similar images in all categories

| | Similarity Metrics | | |
|---|---|---|---|
| Object Category | Quasi-Euclidean Similarity | SSIM | Euclidean Similarity |
| Beach | 0.8293 | 0.4946 | 0.7371 |
| City | 0.8618 | 0.3714 | 0.6957 |
| Dogs | 0.8091 | 0.5532 | 0.7013 |
| Housing | 0.6819 | 0.3267 | 0.5419 |
| People | 0.7822 | 0.6156 | 0.6812 |
| Trees | 0.8067 | 0.1382 | 0.6793 |

Here, SSIM measures the structural similarity between the two input images whereas our proposed method, Quasi-Euclidean calculates the distance between each pixel's intensity value over the full image and the resultant is normalized to express it in terms of a similarity measure.

## 5    Discussion and Conclusion

The evaluation of similarity measures between images as given by different metrics facilitated us to discover that our proposed technique, Quasi-Euclidean measure performs significantly better than Euclidean metric.

It is interesting to note that in some cases the performance yielded by our methodology is a step ahead in recognizing similarity when compared to SSIM i.e evident from similarity measure of trees. This, however, comes as no surprise because tree structures can show wide variation. This is noteworthy in images with a relatively high resolution (in our case all images were of

As a consequence, distance metrics performs significantly better as the pixel intensity of different trees doesn't vary nearly as much as their canopy structure or their overall shape and size does.

The same can be observed in-case of apartments as the varying structure results in a low SSIM score. On the contrary, our technique furnishes a very decent similarity measure. On our novice dataset, Quasi-Euclidean metric almost consistently yielded exceptional results when the images were noticeably similar and even gave expected measures when the images were not similar as projected in Figure 3 through a comparison of metric values for an image of a villa and of an apartment building.

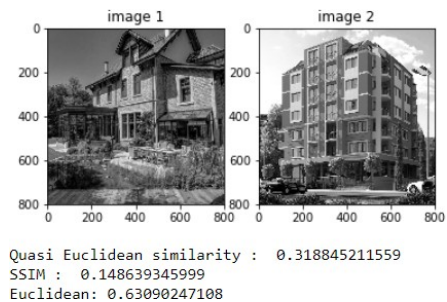

**Fig. 3.** A sample of Similarity measures between an Apartment and a Villa

However, our technique does possess a major limitation which plagues most of the distance metrics. In the images containing completely different objects but having a very similar background or containing different objects but having similar intensity values, our proposed method depicted a very high similarity measure as majority of the pixel values are approximately equal and hence this results in a high similarity value. This is one of the few shortcomings of using pixel intensity values (especially in case of Gray-scale images) for calculating image similarity.

## 6  Future Work

The proposed method works noticeably better than Euclidean Distance metric as well as SSIM in some cases. We also lay emphasis on deploying our technique for the construction of an efficient content-based image retrieval system[7] in the future. The proposed method does have certain limitations, but we perceive that combined with perhaps a method like SSIM or any other method that accounts for structure of the object and optimizing the weights given to the individual methods in the combined system to compute a similarity measure, could potentially make a very efficient and accurate method for an image retrieval system.